\documentclass[10pt,twocolumn,letterpaper]{article}

\usepackage[pagenumbers]{cvpr} % To force page numbers, e.g. for an arXiv version

\usepackage{bm}
\usepackage{amsmath}

\providecommand{\realnum}{\ensuremath{\mathbb{R}}}

\definecolor{cvprblue}{rgb}{0.21,0.49,0.74}
\usepackage[pagebackref,breaklinks,colorlinks,allcolors=cvprblue]{hyperref}

\def\paperID{10575} % *** Enter the Paper ID here
\def\confName{CVPR}
\def\confYear{2025}

\title{AV-Flow: Transforming Text to Audio-Visual Human-like Interactions}

\author{
Aggelina Chatziagapi\\
Stony Brook University\\
{\tt\small aggelina@cs.stonybrook.edu}
\and
Louis-Philippe Morency\\
Meta AI\\
{\tt\small lpmorency@meta.com}
\and
Hongyu Gong\\
Meta AI\\
{\tt\small hygong@meta.com}
\and
Michael Zollh{\"o}fer\\
Codec Avatars Lab, Meta\\
{\tt\small zollhoefer@meta.com}
\and
Dimitris Samaras\\
Stony Brook University\\
{\tt\small samaras@cs.stonybrook.edu}
\and
Alexander Richard\\
Codec Avatars Lab, Meta\\
{\tt\small richardalex@meta.com}
}

\newcommand{\MethodName}{AV-Flow\xspace}

\begin{document}
% \maketitle
\twocolumn[{%
\renewcommand\twocolumn[1][]{#1}%
\maketitle
\vspace{-20pt}
\centering
\includegraphics[width=\linewidth]{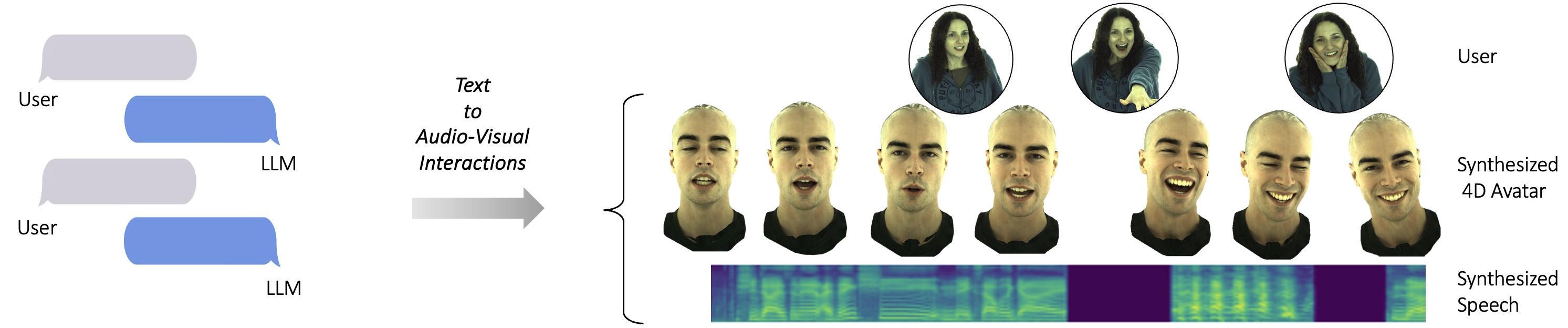}
\captionof{figure}{We present \MethodName, a novel method for \textit{joint audio-visual generation} of 4D talking avatars, given text input only (\eg obtained from an LLM). Inter-connected diffusion transformers ensure cross-modal communication, synthesizing synchronized speech, facial motion, and head motion, based on the flow matching objective. \MethodName further enables empathetic dyadic interactions, by animating an \textit{always-on} avatar that actively listens and reacts to the audio-visual input of a user.}
\label{fig:teaser}
\vspace{8pt}
}]

% Potential titles:

% Fully-Parallel Audio-Visual Generation of 4D Talking Faces via Flow Matching

% Generating Audio-Visual Human-like Interactions via Flow Matching

% Transforming text to audio-visual human-like interactions

% Chatting with a human-like LLM 

% A plug-and-play audio-visual talking head for dyadic interactions

% A plug-and-play audio-visual interface for LLMs

% Audio-Visual Flow Matching for Talking Avatars

% Audio-Visual Flow Matching for Talking Head Generation

% Multi-modal Flow Matching for Talking Head Generation

% Audio-Visual Generation of Talking Avatars via Flow Matching

\begin{abstract}
We introduce AV-Flow, an audio-visual generative model that animates photo-realistic 4D talking avatars given only text input. In contrast to prior work that assumes an existing speech signal, we synthesize speech and vision jointly. We demonstrate human-like speech synthesis, synchronized lip motion, lively facial expressions and head pose; all generated from just text characters. The core premise of our approach lies in the architecture of our two parallel diffusion transformers. Intermediate highway connections ensure communication between the audio and visual modalities, and thus, synchronized speech intonation and facial dynamics (e.g., eyebrow motion). Our model is trained with flow matching, leading to expressive results and fast inference. In case of dyadic conversations,
% we can additionally condition on a participant and synthesize empathetic interactions.
AV-Flow produces an always-on avatar, that actively listens and reacts to the audio-visual input of a user. 
% \MZ{There seem to be multiple use cases, i.e., talking head from text and talking head conditioned on other user. This is unclear until the last sentance of the abstract}
Through extensive experiments, we show that our method outperforms prior work, synthesizing natural-looking 4D talking avatars.
Project page: \small{\url{https://aggelinacha.github.io/AV-Flow/}}.
\end{abstract}

\section{Introduction}
\label{sec:intro}

% {\color{magenta}
With the rise of large language models (LLMs), such as ChatGPT or Llama, we have entered an era where we can communicate and interact with artificial intelligence and knowledge-based systems using human language.
However, such communication is currently largely limited to text only,
% \MZ{limited to verbal social signals?},
with some speech-based exceptions like GPT-4o.
As beings with spatial audio-visual sensing, humans evolved to communicate best through speech and facial expressions.
An immersive and natural interaction between a human and an AI system therefore requires more than just text.

In this work, we aim to get one step closer to closing this gap.
We demonstrate a method that synthesizes a photo-realistic 4D avatar, based on text input only, and generates speech, facial expressions, head motion, and lip sync simultaneously.
Since listening and reactive expressions are just as much part of a communication as speaking, we further show that we can learn active listening behavior -- like back-channeling of expressions -- from the audio-visual input from a user.
Overall, our approach allows to animate an \textit{always-on} 4D avatar by transforming text (\eg, as obtained from an LLM) into expressive speech and facial motion, and exhibits active listening and expressive reactions, leading to empathetic dyadic interactions.
% when it is not speaking.

Our work is closely related to talking face generation, which has been a topic of research exploration for many years.
% Early approaches rely on rule-based methods or hand-engineered features~\cite{?}.
Early works learn phoneme-to-viseme mappings~\cite{bregler1997video,voicepuppetry}.
More recent approaches model talking faces with deep neural networks, conditioned on speech signals, either for 2D videos~\cite{wav2lip,guo2021ad,pcavs,zhang2023sadtalker} or for 3D meshes~\cite{VOCA2019,MeshTalk,fan2022faceformer,aneja2023facetalk}. Generative adversarial networks (GANs)~\cite{goodfellow2014generative} have repeatedly shown appealing results~\cite{wav2lip,Vougioukas_2019_CVPR_Workshops,Yang:2020:MakeItTalk,Jang_2024_CVPR}. Currently, diffusion models~\cite{ho2020denoising} have taken over the generative modeling space, with VASA-1~\cite{xu2024vasa} achieving lifelike generation of audio-driven talking faces in 2D videos.

% One of the first notable works in 2D talking face generation creates convincing lip synchronization of the former U.S.\ President Barack Obama~\cite{?}.
% Generative adversarial networks (GANs) have repeatedly shown appealing results in subsequent works~\cite{?}.
% In the 3D domain, MeshTalk~\cite{?} and FaceFormer~\cite{?} show expressive mesh animation from speech inputs, yet none of them address photorealistic 3D avatars.
% More recently, diffusion models have taken over the generative modeling space and improved talking head models emerged~\cite{?}, including VASA-1~\cite{?}, which achieves lifelike generation of audio-driven talking faces in 2D videos.

However, these works tend to fall short on one or multiple axes.
\textbf{\textit{First}}, most are limited to cascaded systems. They assume that an input speech signal is already provided, either as a real recording of a human voice or generated by a pre-trained text-to-speech system, and mainly focus on precise lip synchronization.
Connecting these systems to an LLM requires a cascaded approach of text-to-speech, followed by speech-to-vision.
%Some propose additional control of facial expressions and/or head pose, conditioning on learned embeddings~\cite{emotional} or other videos~\cite{pc-avs…}.
%Only a few works attempt to drive a talking face from just text, either directly~\cite{text-driven}, or through a cascaded approach of text-to-speech and speech-to-vision~\cite{}.
In contrast, we propose an architecture that \textit{jointly} generates audio and visual outputs, directly from text.
In this way, we achieve natural synchronization of different modalities (\eg, speech intonation and corresponding eyebrow motion) and avoid latency and error accumulation that cascaded systems suffer from.
\textbf{\textit{Second}}, existing systems typically focus on monadic settings.
They can generate facial animation given text or speech, yet do not consider non-speech cases.
In other words, existing systems can actively speak, but they do not generate authentic listening behavior.
We demonstrate that conditioning on audio-visual user input creates authentic listening behavior, such as back-channeling of smiles.
\textbf{\textit{Third}}, most existing works operate on 2D video or on untextured 3D meshes which lack detail.
Our approach operates on high-quality photo-realistic 4D avatars instead.

Our proposed AV-Flow (Audio-Visual Flow Matching) consists of two inter-connected diffusion transformers.
Given input text tokens, one transformer generates the speech signal and the other one	generates the visual output.
%The visual transformer works in the latent space and produces the holistic head and facial dynamics, including lip motion, facial expressions, eyebrow and eyelid movements.
Through intermediate highway connections, we ensure communication between the speech and visual modalities.
% the speech transformer and the visual transformer. 
In this way, AV-Flow synthesizes highly correlated speech and vision (\eg, synchronized speech intonation with facial dynamics).
% (\eg, eyebrow motion).
%It is also versatile in terms of input conditioning: we show that we can drive a talking face given text alone as well as audio-visual user input to improve dyadic conversational behavior.
With additional conditioning on a user's video and audio, we allow our system to reason over the user's behavior and generate avatar motion accordingly, leading to richer expressions, back-channeling of emotional cues like smiles, or affirming head nods.
We resort to flow matching~\cite{lipman2022flow} as a training objective, in order to achieve fast inference and synthesize human-like speech and 4D visual outputs, that capture natural nuances and expressive motion.

In brief, our contributions are as follows:
\begin{itemize}
  % \item We present \MethodName, a method that synthesizes photo-realistic 4D talking avatars only from text.
  \item We introduce AV-Flow, a novel approach for joint audio-visual generation of 4D talking avatars using flow matching, given just input text.
  % \item In contrast to existing speech-driven or cascaded systems, our approach jointly generates speech and facial motion, leading to correlated audio-visual outputs. 
  % style consistency in the audio and visual domain;
  \item Our fully-parallel diffusion transformers ensure cross-modal interaction through intermediate highway connections, generating correlated speech and visual outputs.
  % \item \MethodName is reactive to audio-visual inputs of a user, and therefore enables dyadic conversational scenarios, where the avatar can be \textit{always-on} and actively listens, as opposed to be limited to speaking.
  \item \MethodName enables dyadic conversations, by animating an \textit{always-on} avatar that actively listens and reacts to the audio-visual input of a user.
\end{itemize}

\section{Related Work}
\label{sec:related}

\noindent
\textbf{Audio-driven Talking Heads.}
Earlier approaches for audio-driven talking face generation, such as Video Rewrite~\cite{bregler1997video} and Voice Puppetry~\cite{voicepuppetry}, propose probabilistic models that map phonemes extracted from an audio signal to corresponding mouth shapes (visemes). This phoneme-to-viseme mapping can be learned by hidden Markov models (HMMs)~\cite{sako2000hmm, fu2005audio}, decision trees~\cite{kim2015decision}, or long short-term memory (LSTM) units~\cite{fan2015}. Synthesizing Obama~\cite{suwajanakorn2017synthesizing} is one of the first notable works that produces photo-realistic lip synced videos of former U.S.\ President Barack Obama. Subsequent works propose encoder-decoder architectures, most of them trained as generative adversarial networks (GANs)~\cite{goodfellow2014generative}, using large video datasets
% , in order to achieve appealing visual quality
~\cite{wav2lip,yousaid,Vougioukas_2019_CVPR_Workshops,chen2019hierarchical,zhou2019talking,pcavs,Yang:2020:MakeItTalk,thies2020neural,Lahiri_2021_CVPR}. 
% Wav2Lip~\cite{wav2lip} achieves a highly competitive lip sync accuracy.
Most operate in the 2D space, generating low-resolution videos, while some of them use intermediate representations, like landmarks or 3DMM parameters~\cite{blanz1999morphable,zhang2023sadtalker}. They mainly focus on precise lip synchronization, \eg, Wav2Lip~\cite{wav2lip}. A few enable additional control of head pose~\cite{pcavs,Yang:2020:MakeItTalk} or emotion~\cite{drobyshev2024emoportraits,xu2023high,eamm,wang2022pdfgc,Gan_2023_ICCV}.
Recent approaches produce higher-resolution images, by learning a 3D representation of the human head based on neural radiance fields (NeRFs)~\cite{guo2021ad,semantic,dfanerf,ye2023geneface,ye2023geneface++,lipnerf} or gaussian splatting~\cite{cho2024gaussiantalker,li2025talkinggaussian}. Still, the output is a 2D video and they require a pre-recorded speech signal as input.

Another line of work addresses the problem of audio-driven 4D facial animation. Several works learn to animate subject-specific face models~\cite{karras2017audio,pham2017speech,richard2021audio,cao2005expressive} or artist-designed character rigs~\cite{taylor2017deep,edwards2016jali,zhou2018visemenet} based on input speech. Works~\cite{VOCA2019,aneja2023facetalk,thambiraja2023imitator,xing2023codetalker,Peng_2023_ICCV,sun2024diffposetalk}, like VOCA~\cite{VOCA2019}, MeshTalk~\cite{richard2021meshtalk}, FaceFormer~\cite{fan2022faceformer}, and FaceTalk~\cite{aneja2023facetalk} show expressive animation of 3D meshes, accurately lip syncing to speech inputs. However, they do not learn head motion and only use untextured 3D meshes that lack detail.

% \noindent
% \textbf{Diffusion-based Talking Heads.}
Diffusion models~\cite{ho2020denoising,song2020denoising,song2020score} have recently taken over in the generative modeling domain. They have already shown improved results in talking faces~\cite{stypulkowski2024diffused,shen2023difftalk,ma2023dreamtalk,tian2024emo,xu2024hallo,wei2024aniportrait,kim2024moditalker}. Some of them operate in the image space, while more recent ones propose latent diffusion models; conditioned on audio, they generate latent face embeddings. VASA-1~\cite{xu2024vasa} achieves lifelike generation of audio-driven talking faces. However, all these can only produce 2D videos. More related to our work, Audio2Photoreal~\cite{ng2024audio2photoreal} synthesizes photo-realistic 4D humans. While it generates natural-looking gestures, it lacks in lip syncing, and again assumes existing speech signal as input. In contrast, we propose a method that can animate 4D avatars from just text. Furthermore, we use flow matching~\cite{lipman2022flow}, which compared to diffusion models, achieves faster inference and better performance.

\begin{figure*}[t]
  \centering
   \includegraphics[width=\linewidth]{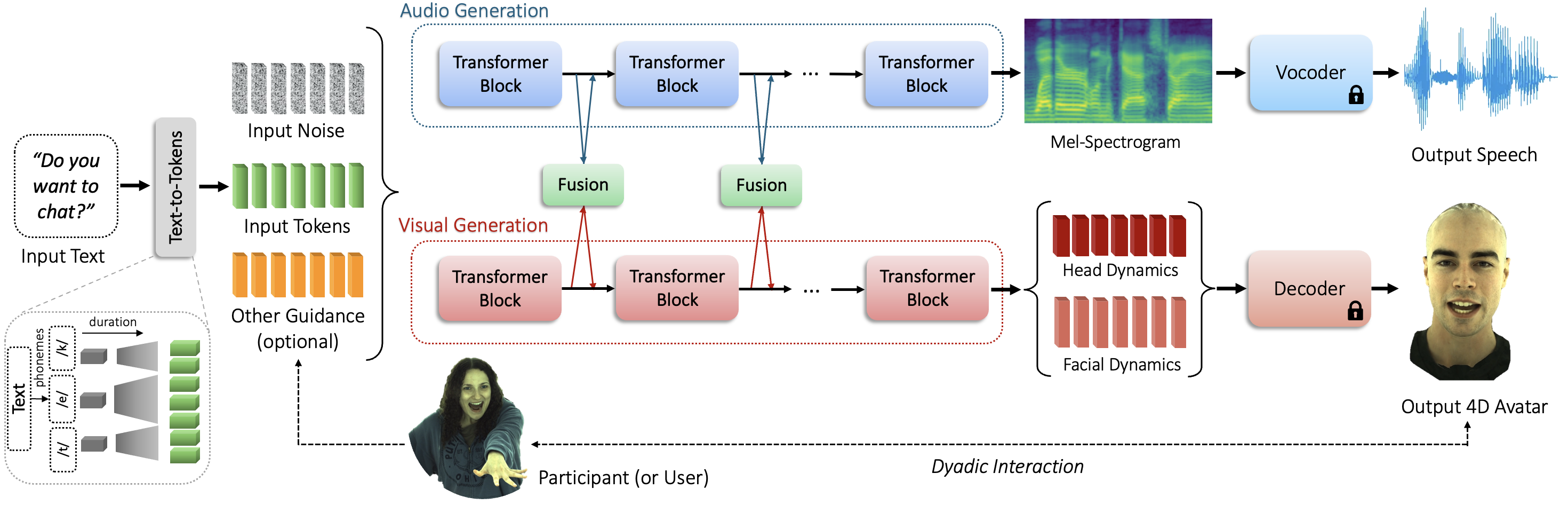}
   \caption{\textbf{Overview of \MethodName}. Given any input text, our method synthesizes \textit{expressive audio-visual 4D talking avatars}, jointly generating head and facial dynamics and the corresponding speech signal. Two parallel diffusion transformers with intermediate highway connections ensure communication between the audio and visual modalities. \MethodName can be additionally conditioned on the audio-visual input of a user, in order to synthesize conversational avatars in dyadic interactions.}
   \label{fig:overview}
   \vspace{-5pt}
\end{figure*}

\noindent
\textbf{Text-driven Talking Heads.} The problem of text-driven talking faces is much less explored. Very early works, like MikeTalk~\cite{ezzat2000visual}, are based on phoneme-to-viseme mappings~\cite{taylor2012dynamic,wan2013photo}. Subsequently, neural networks learn text-to-lip positions in the image space~\cite{kumar2017obamanet,liu2022parallel,choi2024text}, but they work on low-resolution 2D videos. Some works propose cascaded approaches, \ie, text-to-speech (TTS) and then speech or latent codes to vision~\cite{zhang2022text2video,ye2023ada,mitsui2023uniflg}. However, cascaded methods usually are slower and prone to error accumulation. Other works use text for video editing~\cite{fried2019text}, or as a description~\cite{wang2024instructavatar,ling2024posetalk,diao2023ft2tf,xu2023high,ma2023talkclip}, in order to control the emotional state or the identity of the generated subject.
Most related with us are TTSF~\cite{Jang_2024_CVPR} and NEUTART~\cite{milis2023neural}, which jointly generate speech and video. However, they again work with low-resolution 2D videos. In contrast, we synthesize high-quality, photo-realistic 4D avatars. Through our fully-parallel architecture, trained end-to-end with flow matching, we achieve fast inference and natural-looking talking heads. Additionally, our method enables an \textit{always-on} avatar that actively listens and reacts, leading to empathetic interactions with a user.

\section{Method}
\label{sec:method}

We present \MethodName, a novel method for \textit{joint audio-visual generation} of 4D talking avatars, driven by just text inputs. An overview of our approach is illustrated in \cref{fig:overview}. \MethodName consists of two inter-connected diffusion transformers, one for audio and one for visual generation. Given input text tokens, the audio transformer generates a mel-spectrogram, through a series of transformer blocks. Correspondingly, the vision transformer generates head and facial dynamics. We design intermediate highway connections that ensure communication between the audio and visual modalities. The synthesized mel-spectrogram is decoded to a speech signal via a pre-trained vocoder. Correspondingly, the predicted head and facial dynamics are rendered to the output 4D avatar with a pre-trained decoder. The overall method produces audio and visual outputs in a \textit{fully-parallel} way.
% The input tokens can be derived from just text characters or input speech using a speech-to-text model. 
In case of dyadic interactions, we additionally condition on audio-visual signals from a user, which guides the synthesis of natural-looking conversational 4D avatars.

% tokens-to-spectrogram and tokens-to-vision (with interactions) - additional guidance (visual guidance from participant in dyadic conversations)

\subsection{Representations}\label{sec:method_representations}

Our training data consist of dyadic conversations between a main subject, dubbed as ``actor'', and a ``participant'' (or user). The raw audio includes one channel for each of them. The 3D avatar representation of the actor corresponds to the latent space of a Codec Avatar~\cite{deep_appearance, wuu2022multiface}.
% , which we build following~\cite{wuu2022multiface}. 
% This representation is a 256-dimensional vector of latent expression codes for the face and a 6-DoF head rotation and translation.
For the participant, we have access to a monocular video showing their face.
%We have available annotations for the facial dynamics and the head pose of the actor, as well as the raw video of the participant.
We first present our basic \MethodName model, which is trained on the actor's data only (pairs of audio, head and face encodings). In \cref{sec:method_dyadic}, we demonstrate how we can additionally condition on participant's information to model dyadic interactions.

\noindent
\textbf{Input Tokens.}
We transcribe the raw audio using a pre-trained Wav2Vec2 model~\cite{baevski2020wav2vec} for speech recognition (ASR)~\cite{panayotov2015librispeech,yang2021torchaudio}. We extract the outputs of the last layer (logits), which essentially correspond to text character predictions. We use the logits $a_i \in \realnum ^{29}$ per frame $i$ as input tokens of our model.
% (see \cref{fig:overview}).
At inference time, we can drive our model from raw text alone, by learning a text-to-tokens module that maps raw text to character-level logits (see~\cref{sec:method_textdriven}).

% To make representations between training and inference compatible, we train a model that maps from raw text on character level to wav2vec2-style logits, see Section~\ref{?}. \ar{and maybe more detailed description in supp?}

\noindent
\textbf{Facial Dynamics.} Our data include facial expression codes $\bm{f}_i \in \realnum ^ {256}$ per frame $i$ of the actor, which correspond to the latent space of a VAE~\cite{deep_appearance}. These represent the holistic facial motion, including facial expression, lip motion, eyebrow and eyelid movements.

\noindent
\textbf{Head Dynamics.} 
% Our data also include head poses, extracted as a rotation matrix and translation vector.
Our data also include a 6-DoF head rotation and translation. We convert the rotation matrix to quarternion representation and concatenate with the translation vector, leading to a head pose $\hat{\bm{h}}_{i} \in \realnum ^ {7}$ per frame $i$. We train a small temporal VAE on the head poses, with one transformer encoder layer for the encoder and one for the decoder, in order to learn a more robust and temporally consistent representation. We use the latent space of this VAE to encode the head dynamics $\bm{h}_{i} \in \realnum ^ {8}$ per frame $i$.

% Expression codes 

% Head pose (quartenion and temporal VAE)

% Participant (FLAME expression codes and head pose from SMIRK) - Dyadic setting

\subsection{Architecture}\label{sec:method_architecture}

\noindent
\textbf{Diffusion Transformers.} We construct 2 parallel diffusion transformers, one for the audio generation and one for the visual generation. Each of them is based on the original architecture of latent diffusion transformer (DiT) models~\cite{peebles2023scalable}, and consists of $N$ blocks. We follow the variant of \textit{in-context conditioning}, concatenating input noise with the input tokens. The transformers are trained to progressively denoise the input noise, in order to restore the corresponding signal, and model the appropriate distribution. Formally, for a sequence of $n$ frames with input tokens $A = \{\bm{a}_1, \bm{a}_2, \dots, \bm{a}_n\}$, the audio transformer learns to synthesize the corresponding mel-spectrogram $S \in \realnum ^{n \times 80}$, and the visual transformer learns to synthesize the corresponding
head poses $H = \{\bm{h}_1, \bm{h}_2, \dots, \bm{h}_n\}$ and facial encodings $F = \{\bm{f}_1, \bm{f}_2, \dots, \bm{f}_n\}$. In order to facilitate the communication between the modalities and achieve exact correspondence of the number of frames and spectrogram dimension, we upsample the ground truth annotations to the rate of the spectrogram bins (at 86 fps).

\noindent
\textbf{Audio-Visual Fusion.} We design intermediate highway connections that enable communication between the audio and visual modalities. In this way, we achieve synchronized speech intonation with facial and head dynamics. Formally, for an output $\bm{x}^{a}_{l} \in \realnum ^{d_l}$ of a DiT block $l$ of the audio transformer ($l = 1, \dots, N$), and the corresponding output $\bm{x}^{v}_{l}  \in \realnum ^{d_l}$ of the vision transformer, we learn a linear fusion:
\begin{align}\label{eq:fusion}
    \bm{y}^{a}_{l} = \bm{x}^{a}_{l} + \bm{U}_{l}^\intercal [\bm{x}^{a}_{l}; \bm{x}^{v}_{l}] + \bm{b}_{l}, \\
    \bm{y}^{v}_{l} = \bm{x}^{v}_{l} + \bm{V}_{l}^\intercal [\bm{x}^{a}_{l}; \bm{x}^{v}_{l}] + \bm{c}_{l},
\end{align}
where $\bm{U}_{l}, \bm{V}_l \in \realnum ^{2d_l \times d_l}$ and $\bm{b}_{l}, \bm{c}_{l} \in \realnum ^{d_l}$ are learnable parameters.
The resulting features $ \bm{y}^{a}_{l} $ and $ \bm{y}^{v}_{l} $ are then fed as input into the next transformer block $ l+1 $ of the audio and visual transformer, respectively (see~\cref{fig:overview}).

\noindent
\textbf{Vocoder.} We decode the synthesized mel-spectrogram to a speech signal using a pre-trained BigVGAN vocoder~\cite{lee2022bigvgan}. We use the base version that is trained with speech signals sampled at 22050 Hz and spectrograms with 80 mel bands. We keep its weights frozen.

\noindent
\textbf{Avatar Decoder and Renderer.} We use the mesh-based Codec Avatar decoder and renderer released in~\cite{ng2024audio2photoreal,timur2021driving_renderer}.

% Diffusion Transformer

% Audio-to-vision module

\subsection{Flow Matching}

Flow matching is an efficient approach for generative modeling, recently introduced by Lipman \etal~\cite{lipman2022flow}. It combines ideas from continuous normalizing flows (CNF) and diffusion models. It leads to simpler paths with straight line trajectories, compared to the curved paths of diffusion models. Thus, it enables faster training and inference. In this section, we describe an overview of flow matching, that we use to train \MethodName. We refer the interested reader to~\cite{lipman2022flow}.

Let $\bm{x} \in \realnum ^{d}$ an observation in the data space (that can be a spectrogram sample or head pose or face encoding in our case), sampled from an unknown distribution $q(\bm{x})$. A probability density path is a time-dependent probability density function $p_t: [0, 1] \times \realnum ^d \rightarrow \realnum ^ {+} $. CNFs construct a probability path $p_t$ such that $p_0$ is a simple prior distribution, \ie, a standard normal distribution $p_0(\bm{x}) = \mathcal{N}(\bm{x}; \bm{0}, \bm{I})$, and $p_1$ approximates the distribution $q$. A time-dependent vector field $u_t: [0, 1] \times \realnum ^d \rightarrow \realnum ^ {d}$ generates the path $p_t$, and is used to construct a flow $\phi_t: [0, 1] \times \realnum ^d \rightarrow \realnum ^ {d}$; $\phi_t$ pushes the data from the prior towards the target distribution and is defined via the ordinary differential equation (ODE):
\begin{equation}
    \frac{d}{dt} \phi_t (\bm{x}) = v_t(\phi_t(\bm{x})) ; \quad \phi_0(\bm{x}) = \bm{x}.
    \label{eq:ode}
\end{equation}
The vector field $v_t$ is approximated by a neural network with parameters $\bm{\theta}$. Flow matching proposes the following objective, that allow us to flow from $p_0$ to $p_1$:
\begin{equation}\label{eq:fm_objective}
    \mathcal{L}_{\text{FM}}(\bm{\theta}) = \mathop{\mathbb{E}_{t, p_t(\bm{x})}} \| v_t(\bm{x}; \bm{\theta}) - u_t(\bm{x}) \|^2.
\end{equation}
As a tractable instantiation of Eq.~\eqref{eq:fm_objective}, we follow the Optimal Transport (OT) formulation from~\cite{lipman2022flow} where the flow from $p_0$ to $p_1$ is modeled by a straight line.
Then,
\begin{align}
    \phi_t(\bm{x}) = (1 - (1-\sigma_\text{min})t) \bm{x} + t\bm{x}_1
\end{align}
and the network $ v_t $ predicting the flow from a random Gaussian sample $\bm{x}_0 \sim p_0(\bm{x})$ to a data sample $\bm{x}_1 \sim q(\bm{x}_1)$ can be trained by optimizing the conditional flow matching objective:
\begin{align}\label{eq:flowloss}
    \mathcal{L}_{\text{CFM}} = \mathbb{E}_{t, \bm{x}_0, \bm{x}_1} \| v_t(\phi_t(\bm{x}_0)) - \big(\bm{x}_1 - (1 - \sigma_\text{min})\bm{x}_0\big) \|^2.
\end{align}
We empirically found that an L1 loss leads to more realistic results than an L2 loss.
Therefore, we optimize the objective:
\begin{equation}\label{eq:ourloss}
    \mathcal{L}_{\text{AV-Flow}} = \lambda _{s} \mathcal{L}_{s} + \lambda _{h} \mathcal{L}_{h} + \lambda _{f} \mathcal{L}_{f} \; ,
\end{equation}
where $\mathcal{L}_{s}$, $\mathcal{L}_{h}$, $\mathcal{L}_{f}$ are the objectives as in Eq.~\eqref{eq:flowloss} for the mel-spectrograms $S$, head poses $H$, and facial dynamics $F$ correspondingly, using L1 norm instead of L2.
Once the network $ v_t $ is trained, any ODE solver can be used to solve Eq.~\eqref{eq:ode}. We use the Euler solver in our work.

\subsection{Text-Driven Generation}\label{sec:method_textdriven}

As mentioned in \cref{sec:method_representations}, during training, our input tokens are logits extracted from an ASR model, since we do not have available any text annotations. These input tokens are essentially predictions of text characters. Thus, our model easily generalizes to any input text at inference. To demonstrate this capability, we train a small text-to-tokens model (see \cref{fig:overview}), that maps raw text to logits, using the LJSpeech dataset~\cite{ljspeech17}. It follows a similar architecture with~\cite{mehta2024matcha}.
% , but predicts logits (not spectrograms).
It first maps the input text to phonemes and learns corresponding embeddings. It then predicts their duration and projects them to character-level logits at 86 fps (see also suppl.).
% (see more details in the suppl.~document).
% \ar{this needs more details. It's unclear how we go from raw text to 86fps logits/characters. It's even unclear \textit{that} we go from raw text to 86fps logits/characters. Maybe refer to supp for details and explain it there to save the space here?}

% Audio-Visual Generation

% tokens from ASR: tokens-to-vision + tokens-to-speech

% Interaction between Modalities (Fusion)

% Vocoder

\subsection{Dyadic Conversations}\label{sec:method_dyadic}

% dyadic setting - visual guidance from participant

An important capability of AV-Flow is that it can be easily conditioned to other input signals, that can guide the audio-visual generation accordingly. Using our conversational data (described in more detail in \cref{sec:experiments}), we demonstrate how we can guide the 4D talking avatar in a dyadic interaction, based on the audio-visual input of a user (see \cref{fig:overview}). 
In this way, we synthesize a \textit{conversational} avatar, that \textit{actively listens} and reacts (\eg, with facial expressions or head nodding), leading to empathetic interactions.
% \ar{One important point to highlight is that this allows us to model active listening, where the avatar reacts to the user even when it's not speaking.}

We propose to provide audio and visual guidance from the raw monocular video of the participant. Given each video frame $i$, we extract features $\bm{s}_i$ using SMIRK~\cite{SMIRK:CVPR:2024}. SMIRK predicts FLAME~\cite{FLAME:SiggraphAsia2017} parameters given a single image and faithfully captures a large variety of facial expressions. The features $\bm{s}_i = [\bm{e}_i; \bm{j}_i; \bm{r}_i]$ include facial expressions $\bm{e}_i \in \realnum ^{50}$, jaw pose $\bm{j}_i \in \realnum ^{3}$, and head rotation $\bm{r}_i \in \realnum ^{3}$ per frame $i$. We further extract ASR tokens $\bm{a}_i^{p}$ from the audio channel of the user, similarly with our input tokens for the actor. Both $\bm{s}_i$ and $\bm{a}_i^{p}$ are concatenated with the input tokens, giving audio-visual guidance to our method and enabling dyadic interaction with a user.

% We also evaluate the case where we do not have available the video of the participant, but only the speech signal. In this case, we can extract ASR tokens $\bm{a}_i^{p}$ from the corresponding audio channel, similarly with our input tokens for the actor.

% \section{Final copy}

% You must include your signed IEEE copyright release form when you submit your finished paper.
% We MUST have this form before your paper can be published in the proceedings.

% Please direct any questions to the production editor in charge of these proceedings at the IEEE Computer Society Press:
% \url{https://www.computer.org/about/contact}.
\section{Experiments}
\label{sec:experiments}

\begin{figure*}[t]
  \centering
   \includegraphics[width=0.52\linewidth]{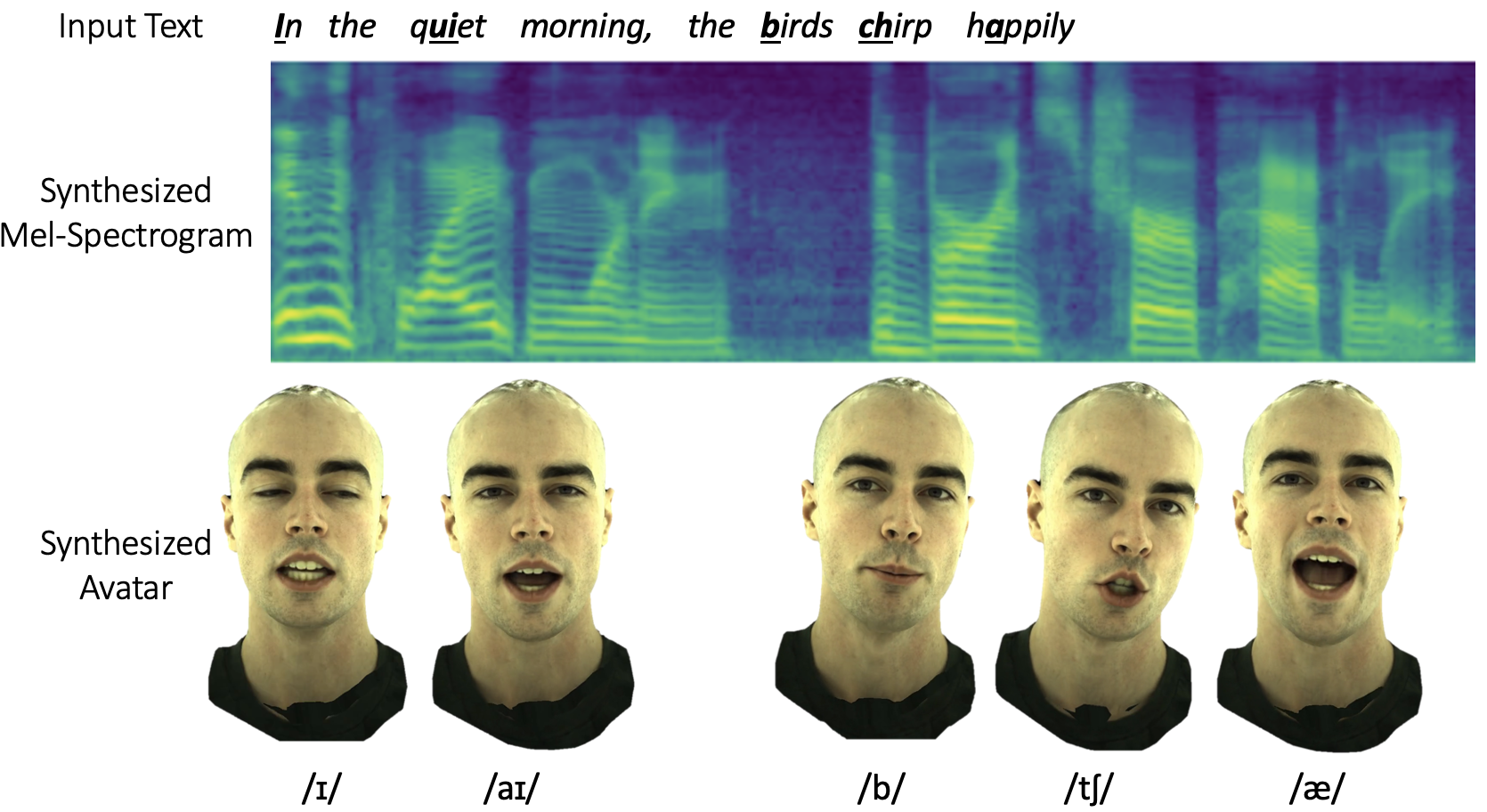}
   \includegraphics[width=0.46\linewidth]{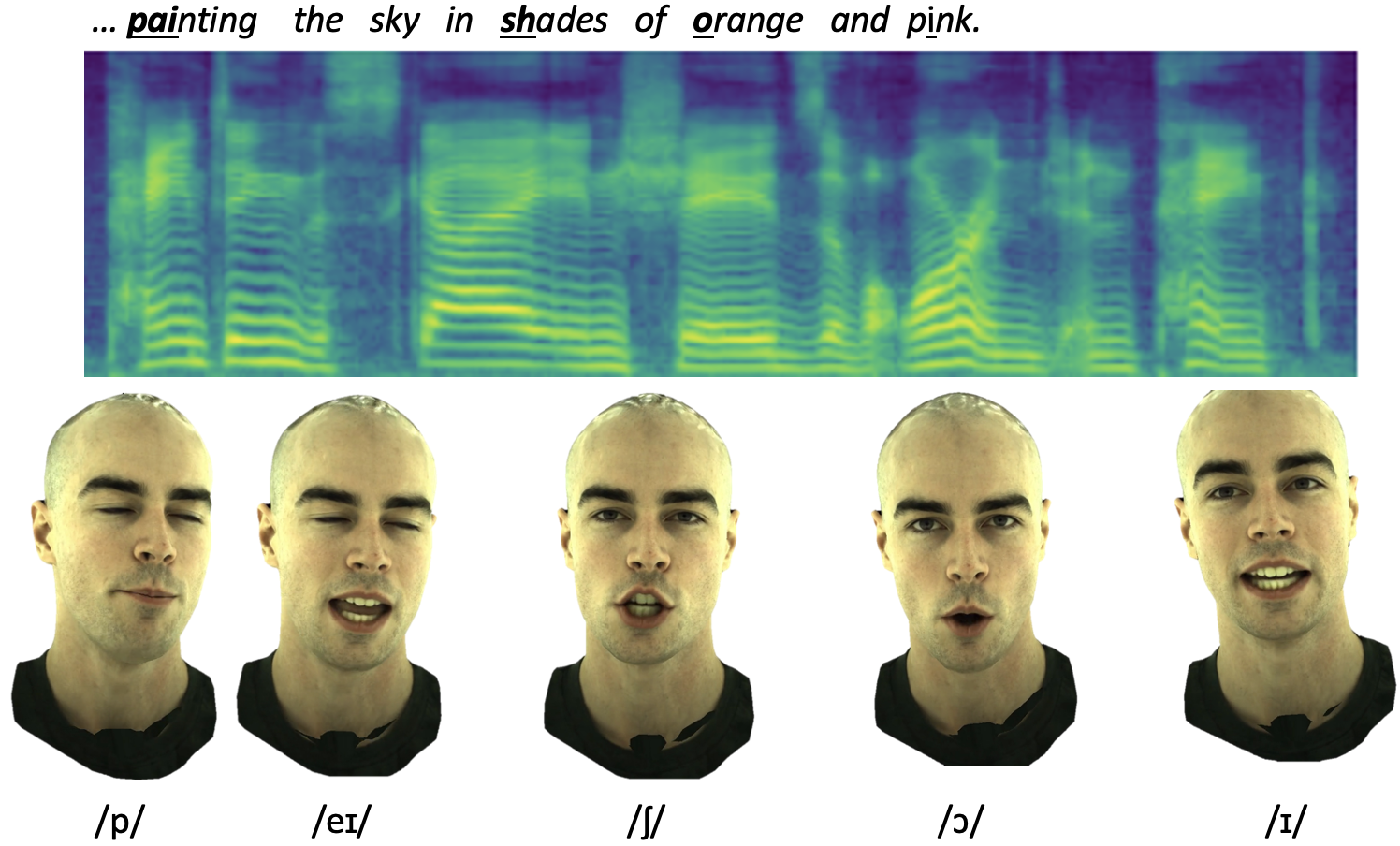}
   \vspace{-5pt}
   \caption{\textbf{Qualitative Results of \MethodName}. From just raw text characters as input, \MethodName synthesizes expressive audio signal (shown as mel-spectrogram on top) and corresponding head and facial dynamics of our 4D talking avatar.}
   \label{fig:qual_text_melspec_avatar}
   % \vspace{-5pt}
\end{figure*}

\begin{table*}[t]
  \centering
  \begin{tabular}{@{}l|ccccccccc@{}}
  \toprule
    & \textit{Lip Sync} & \multicolumn{1}{c}{\textit{Realism}} & \multicolumn{2}{c}{\textit{Diversity}} & \multicolumn{2}{c}{\textit{AV-Alignment}} & \multicolumn{2}{c}{\textit{Audio Quality}} \\
    \midrule
    Method & F1$_{lips}$$\uparrow$ & FD$_{e}$$\downarrow$ & Div$_{h}$$\uparrow$ & Div$_{e}$$\uparrow$ & BC$_{h}$$\uparrow$ & BC$_{e}$$\uparrow$ & MCD$\downarrow$ & WER$\downarrow$ \\
    \midrule
    Separate Models & 0.933 & 0.981 & 0.023 & 0.614 & 0.218 & 0.184 & 1.009 & 0.179 \\
    Shared Weights & 0.910 & 0.862 & 0.024 & 0.664 & 0.208 & 0.174 & 0.986 & 0.287\\
    Cascaded & 0.848 & 1.223 & 0.026 & 0.571 & 0.222 & 0.222 & 1.009 & 0.179 \\
    \midrule
    % AV-Flow w/ Fusion (b) & 0.865 & 0.001 & \textbf{0.828} &  0.022 & \textbf{0.695} & \textbf{0.271} & 0.186 & 0.880 & \\
    AV-Flow (Ours) & \textbf{0.964} & \textbf{0.861} & \textbf{0.029} & \textbf{0.680} & \textbf{0.258} & \textbf{0.229} & \textbf{0.900} & \textbf{0.157}\\
    \bottomrule
  \end{tabular}
  \caption{\textbf{Ablation Study.} We compare with the following variants: (a) Separate Models: 2 separate DiTs (one for audio and one for visual generation), without any connections, (b) Shared Weights: 1 model for both modalities with shared weights, (c) Cascaded Method: sequence of audio DiT (tokens-to-speech) and visual DiT (speech-to-video). Our proposed \MethodName achieves the best results.}
  \label{tab:quant_ablation}
  \vspace{-10pt}
\end{table*}

\noindent
\textbf{Datasets.} We use the publicly available dataset proposed by Audio2Photoreal~\cite{ng2024audio2photoreal}. This dataset includes dyadic conversations between 4 pairs of subjects. Each session lasts about 2 hours, with a total duration of 8 hours. There are 4 individuals. In each session one is the main ``actor'' and the other one is the ``participant''.  The actors are prompted to a diversity of situations, including informal and more professional interactions. Both subjects are captured simultaneously in multi-view capture domes, enabling photo-realistic rendering. The data include the raw audio, face expression codes of the actors, and pre-trained personalized renderers.

In addition to this dataset, we capture an internal dataset of 50 hours in a similar setting. It includes 1 main actor, who is engaged in dyadic conversations with 20 different participants. Similarly with~\cite{ng2024audio2photoreal}, we extract the raw audio at 48 kHz. Applying a simple voice activity detection (VAD), we separate the audio of the actor from the audio of the participant. We extract face encodings and head poses of the full 50 hours for the actor. We also have the raw video of the participant from one camera view and a pre-trained personalized renderer for the actor.

% \noindent
% \textbf{Baselines.}

\noindent
\textbf{Evaluation Metrics.} Since our approach synthesizes both audio and vision, we evaluate both modalities. Regarding the visual part, we follow similar works~\cite{ng2024audio2photoreal,richard2021meshtalk,richard2021audio} and choose a combination of metrics that capture:
% the lip synchronization, the realism, and the diversity of the synthesized avatars:
\begin{itemize}
    \item \textit{Lip synchronization}: We first reconstruct the ground truth and generated 3D meshes per frame. We determine the lip closures when the corresponding vertices of the inner upper and lower lips match (\ie, the distance is almost zero). We compute the F1-score to emphasize the importance of both high precision and high recall~\cite{richard2021audio}.
    \item \textit{Realism}: We measure the Fr\'echet distance between ground truth and generated face expressions (FD$_{e}$), in order to estimate the distribution distance.
    \item \textit{Diversity}: We measure the diversity of the generated head poses (Div$_{h}$) and expressions (Div$_{e}$) as the standard deviation across samples in our test set. 
\end{itemize}
An important contribution of our method is the correlation between our synthesized audio and video:
\begin{itemize}
    \item \textit{Audio-visual alignment}: We calculate the Beat Align Score~\cite{zhang2023sadtalker,zhu2023taming,siyao2022bailando,li2021ai} between audio and head motion (BC$_{h}$) and between audio and facial motion (BC$_{e}$). The audio beats are estimated by detecting peaks in onset strength~\cite{ellis2007beat,mcfee2015librosa} in the generated speech, and the motion beats as the local minima of the kinetic velocity~\cite{li2021ai}.
\end{itemize}
Finally, we evaluate our synthesized speech signal:
\begin{itemize}
    \item \textit{Audio quality}: We measure the mel cepstral distortion (MCD), which estimates the difference between mel cepstra with dynamic time warping~\cite{lee2022bigvgan,Jang_2024_CVPR}, and the word error rate (WER) to evaluate the intelligibility~\cite{paszke2019pytorch,yang2021torchaudio}.
    % , using a pre-trained Wav2Vec2 model for ASR~\cite{paszke2019pytorch}.
\end{itemize}

\noindent
\textbf{Implementation Details.}
In ~\cref{eq:ourloss}, we use $\lambda_s = 3.0$, $\lambda_f = 1.0$, and $\lambda_h = 0.2$. We notice that 8 steps for the Euler solver give similar results with 16 or 32 steps, and lead to faster inference. We use rotary positional embeddings~\cite{su2024roformer} and windows of 10 frames for the DiTs. For our input data rate of 86 fps (see~\cref{sec:method_architecture}), this leads to a negligible starting latency of around 120 ms, and real-time synthesis.
%See suppl.~for more details.

% \textbf{Metrics}

% Vision generation: lip closures, FID, diversity in expressions/head poses

% Speech generation: Audio quality, WER (if text available)

% Beat alignment (Beat Align Score or Beat Consistency Score) similar with other works~\cite{zhang2023sadtalker,zhu2023taming,siyao2022bailando}, where audio beats are estimated by detecting peaks in onset strength~\cite{ellis2007beat}.
% % librosa.beat.beat\_track to calculate the beats

% Diversity: std of expressions and head poses.

\subsection{Ablation Study}\label{sec:exp_ablation}

We first conduct an ablation study, comparing our proposed audio-visual fusion with different variants: (a) Separate models: we train 2 separate transformers in parallel (one for audio and one for visual generation), without any connections. (b) Shared weights: we train a single model for both modalities by concatenating the inputs/outputs. (c) Cascaded approach: we learn the audio DiT (tokens-to-speech) followed by the visual DiT (speech-to-video). \cref{tab:quant_ablation} shows the corresponding quantitative results. \MethodName achieves the best results across all the metrics. It produces well-synchronized lips, with accurate lip closures, which is crucial for photo-realistic 4D talking avatars. In addition, the proposed audio-visual fusion leads to the best correlation between audio and motion, as measured by the BC$_{h}$ and BC$_{e}$ metrics.

\cref{fig:qual_text_melspec_avatar} shows qualitative results of our method. From just raw text characters as input (no audio available), \MethodName synthesizes expressive audio signal and corresponding facial and head dynamics of our 4D talking avatar. Notice the accuracy of the lip motion for each phoneme (written at the bottom), as well as the expressiveness of the avatar.

\begin{figure*}[t]
  \centering
   \includegraphics[width=\linewidth]{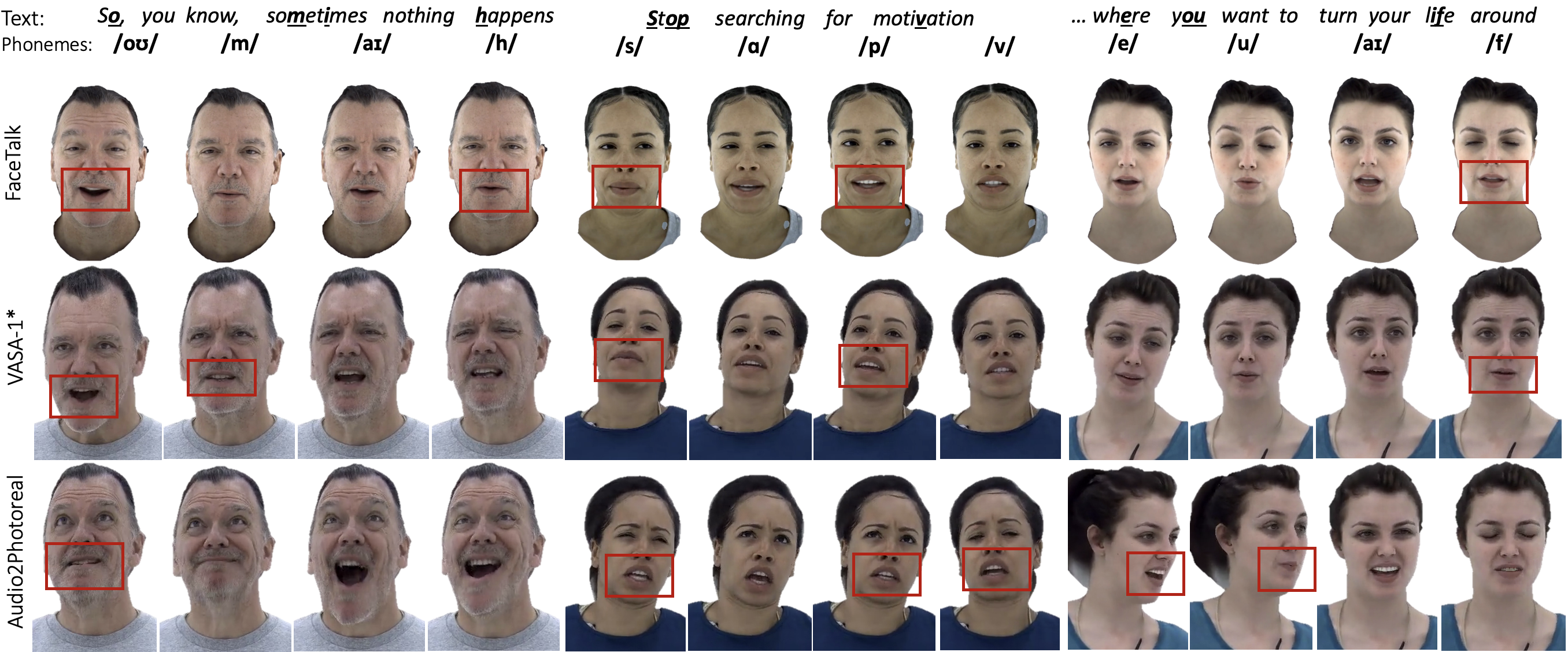}
   \includegraphics[width=\linewidth]{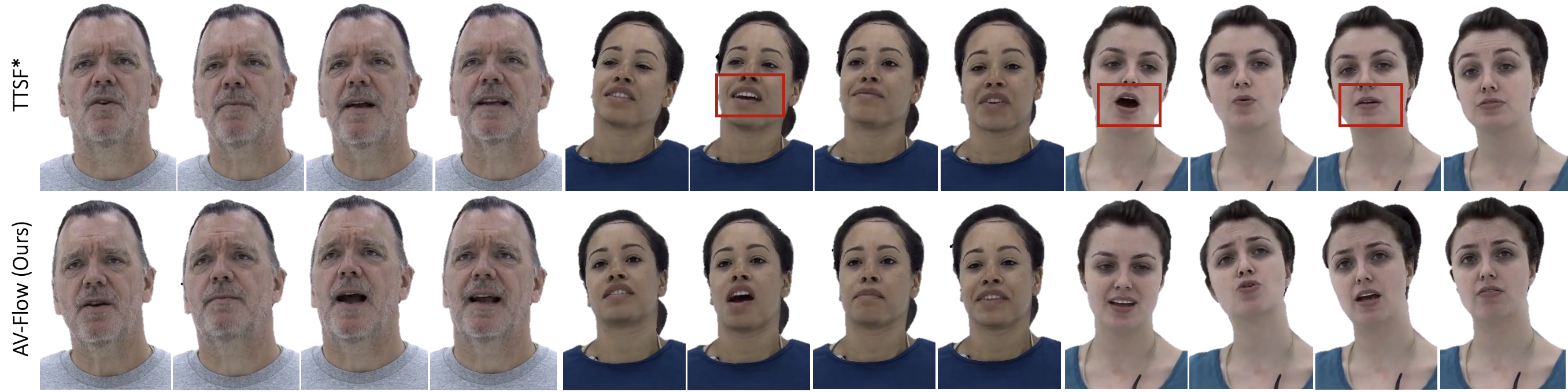}
   \caption{\textbf{Qualitative Evaluation.} We compare with state-of-the-art methods for audio-driven talking faces, namely FaceTalk~\cite{aneja2023facetalk}, VASA-1~\cite{xu2024vasa}, Audio2Photoreal~\cite{ng2024audio2photoreal}, and the text-driven TTSF~\cite{Jang_2024_CVPR} (the only one that can generate speech from text like ours). We re-implement VASA-1 and TTSF (denoted with an asterisk) for our data (face encodings and renderers). FaceTalk only animates the face (not head motion). Our proposed \MethodName synthesizes the corresponding phoneme (shown on top) more accurately.}
   \label{fig:qual_lipsync}
   % \vspace{-5pt}
\end{figure*}

\begin{table*}
  \centering
  \begin{tabular}{@{}l|ccccccccc@{}}
    \toprule
    % Method & Lips & FID/FVD & Diversity - Exp / Head & Alignment with Speech - Exp/Head &  Audio Quality & Speed \\
    & \textit{Lip Sync} & \multicolumn{1}{c}{\textit{Realism}} & \multicolumn{2}{c}{\textit{Diversity}} & \multicolumn{2}{c}{\textit{AV-Alignment}} & \multicolumn{2}{c}{\textit{Audio Quality}} & {\textit{Speed}} \\
    \midrule
    Method & F1$_{lips}$$\uparrow$ & FD$_{e}$$\downarrow$ & Div$_{h}$$\uparrow$ & Div$_{e}$$\uparrow$ & BC$_{h}$$\uparrow$ & BC$_{e}$$\uparrow$ & MCD$\downarrow$ & WER$\downarrow$ & Time (s)$\downarrow$ \\
    \midrule
    % Another audio-driven &&&&& \\
    FaceTalk & 0.851 & 0.873 & N/A & 0.670 & N/A & 0.209 & N/A & N/A & 1.443 \\
    % SadTalker & \\
    VASA-1$^{*}$ & 0.846 & 0.887 & 0.032 & 0.664 & 0.210 & 0.204 & N/A & N/A & 0.965 \\
    Audio2Photoreal & 0.920 & 0.879 & 0.022 & 0.586 & 0.198 & 0.202 & N/A & N/A & 1.578 \\
    \midrule
    VASA-1$^{*}$ w/ TTS & 0.710 & 2.665 & \textbf{0.033} & 0.678 & 0.190 & 0.201 & N/A & N/A & 1.265 \\
    Audio2Photoreal w/ TTS & 0.813 & 3.289 & 0.021 & 0.538 & 0.168 & 0.175 & N/A & N/A & 1.778\\
    TTSF$^{*}$ & 0.929 & 0.962 & 0.023 & 0.630 & 0.233 & 0.211 & 1.229 & 0.285 & 0.400  \\
    \midrule
    AV-Flow (Ours) & \textbf{0.964} & \textbf{0.861} & 0.029 & \textbf{0.680} & \textbf{0.258} & \textbf{0.229} & \textbf{0.900} & \textbf{0.157} & \textbf{0.398} \\
    \bottomrule
  \end{tabular}
  \caption{\textbf{Quantitative Evaluation.} We compare with state-of-the-art methods for audio-driven talking faces, namely FaceTalk~\cite{aneja2023facetalk}, VASA-1~\cite{xu2024vasa}, and Audio2Photoreal~\cite{ng2024audio2photoreal}. We convert them to text-driven by attaching our TTS (denoted w/ TTS), and compare with TTSF~\cite{Jang_2024_CVPR} which is the only method that can generate speech from text like ours. We re-implement VASA-1 and TTSF (denoted with an asterisk) for our 3D data.  We evaluate the synthesis quality, as well as the inference speed (seconds for generating 20 sec.~offline).}
  \label{tab:quant_results}
  \vspace{-10pt}
\end{table*}

\subsection{Evaluation}

\textbf{Baselines.} Very few works for talking faces have addressed the problem of text-driven generation, and even fewer the problem of joint audio-visual generation. To the best of our knowledge, our proposed \MethodName is the first approach that can generate audio-visual 4D talking heads from only text input. A concurrent work with us is TTSF~\cite{Jang_2024_CVPR}. However, we identify the following  main differences: (a) TTSF only generates 2D videos (not 4D avatars). (b) We propose a fully-parallel architecture with intermediate highway connections, focusing on the audio-visual fusion. (c) Our architecture is trained end-to-end with flow matching, achieving fast inference, compared with the GAN-based TTSF. (d) We additionally address the case of dyadic conversations, where our model communicates with a user.  Since TTSF's code is not available, we implement their method for our data (TTSF$^{*}$), where we train a personalized model that predicts face encodings and head motion (therefore, without any GAN-based losses and identity prediction). 

Most of the state-of-the-art methods are audio-driven and focus on the generation of 2D talking faces. We choose to compare with the seminal VASA-1~\cite{xu2024vasa} that similarly with us trains a DiT (but not with flow matching) on the latent space of head and facial dynamics. We adapt VASA-1 to be able to deal with our 3D data (encodings and renderer) and name the variant VASA-1$^{*}$. We also compare with the audio-driven face generation of Audio2Photoreal~\cite{wan2013photo}, as well as FaceTalk~\cite{aneja2023facetalk}. Note that FaceTalk only generates expressions, not head motion, but focuses on 4D avatars compared with the other 2D methods. Finally, we attach our text-to-speech (TTS) model and convert VASA-1$^{*}$ and Audio2Photoreal to text-driven methods. For fair comparison with all these methods, in our evaluation we use input audio from our test set that is converted to audio features (audio-driven) or text tokens (text-driven). 

\noindent
\textbf{Quantitative Evaluation.} \cref{tab:quant_results} demonstrates the corresponding quantitative results. \MethodName produces the best audio-visual alignment, as well as the most accurate lip synchronization. It also synthesizes diverse face and head motion. VASA-1$^{*}$ with TTS seems to slightly surpass our method in the head diversity (Div$_{h}$). However, qualitatively we noticed that this is because it generates more noisy head motion (higher variance). Our input tokens are also more robust as input: in the case of our conversational data, VASA-1$^{*}$ generates noisy and random motion during pauses of the actor. Additionally, we produce better audio quality compared with TTSF~\cite{Jang_2024_CVPR}. 

\noindent
\textbf{Inference Speed.} Since our method is based on flow-matching end-to-end, it achieves faster inference than the other methods. We measure the inference speed as the time for a single pass of the model on a single A100 GPU, averaged across multiple runs. We assume audio features stored
% (otherwise Wav2Vec2 feature extraction would take another 3-4 seconds for FaceTalk and VASA-1)
and omit the renderer for this calculation. The last column of \cref{tab:quant_results} shows this time in seconds. Using only 8 steps for the Euler solver, \MethodName needs only around 200ms to synthesize around 20 seconds of audio-visual content (offline speed). With the additional text-to-tokens module, it requires less than 400ms. Our TTSF$^{*}$ does not include any identity prediction or StyleGAN-based architecture, and thus the inference becomes faster from the original TTSF~\cite{Jang_2024_CVPR}. In comparison, the text-driven VASA-1$^{*}$ that is diffusion-based needs more than 1 second for the same length of video generation.

\noindent
\textbf{Qualitative Evaluation.} 
\cref{fig:qual_lipsync} shows qualitative comparisons with state-of-the art methods for talking face generation. 
% Here, we collect test audio from internet videos and
We use test audio from the EARS dataset~\cite{richter2024ears}, and convert it to input audio features or text tokens accordingly. Since this input comes from a different individual than the training subject, lip syncing becomes more challenging. We show a variety of phonemes (on top) and corresponding mouth positions for each method. \MethodName demonstrates significant robustness, expressively and accurately animating the 4D talking avatars, under any input text. We also encourage the readers to watch our supplemental video.

% Variants:

% - 1 model (text-to-vision)

% - 2 separate models (text-to-vision and text-to-speech): voice quality degrades, no alignment between expressivness in speech and visual result

% - 2 models with shared weights = 1 model with 2 outputs

% - 2 models with interactions (linear fusion after each transformer block)

% - cascaded model (text-to-speech --> speech-to-vision)

% Fusion variants (supplementary):

% - linear fusion, linear fusion with skip connection, beta-gamma from AdaIn

% Compare audio quality and WER with TTS methods (e.g. Matcha-TTS) - can be a small table in supplementary

% TTS + audio-driven method (cascaded) --> higher inference speech (ours fully parallel)

% 60fps offline speed (vs 45fps vasa-1) on A100 (300ms + 800ms + 8.5s = 10sec for 600 frames at 30 fps)
% with nfe steps = 8 --> 198ms, with 4 --> 82ms (for 1725 frames)
% adding 200ms for TTS

% 1.45 for logmel, 4.92 for wav2vec

\begin{figure}[t]
  \centering
   \includegraphics[width=\linewidth]{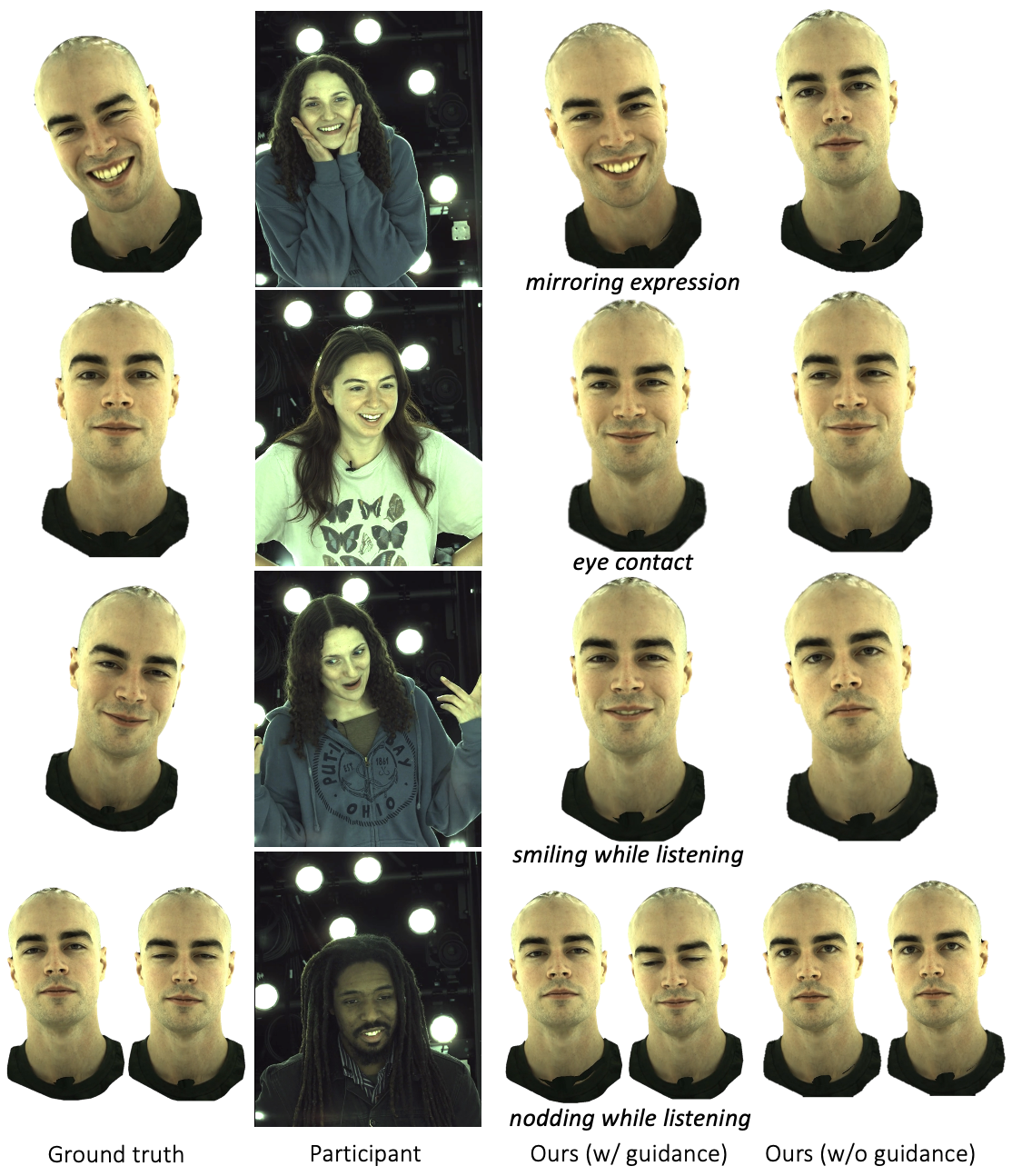}
   % \includegraphics[width=\linewidth]{images/smirk2.png}
   % \vspace{-5pt}
   \caption{\textbf{Audio-Visual Guidance in Dyadic Conversations.} The actor reacts (with their gaze or smile) according to the participant's expression and/or voice (\MethodName with guidance).}
   \label{fig:qual_smirk}
   \vspace{-10pt}
\end{figure}

% \begin{figure}[t]
%   \centering
%    \includegraphics[width=\linewidth]{images/headexp_spec.png}
%    \caption{Correlation between synthesized speech, head pose and facial dynamics.}
%    \label{fig:spec_head_exp}
% \end{figure}

\begin{figure}[t]
  \centering
   \includegraphics[width=\linewidth]{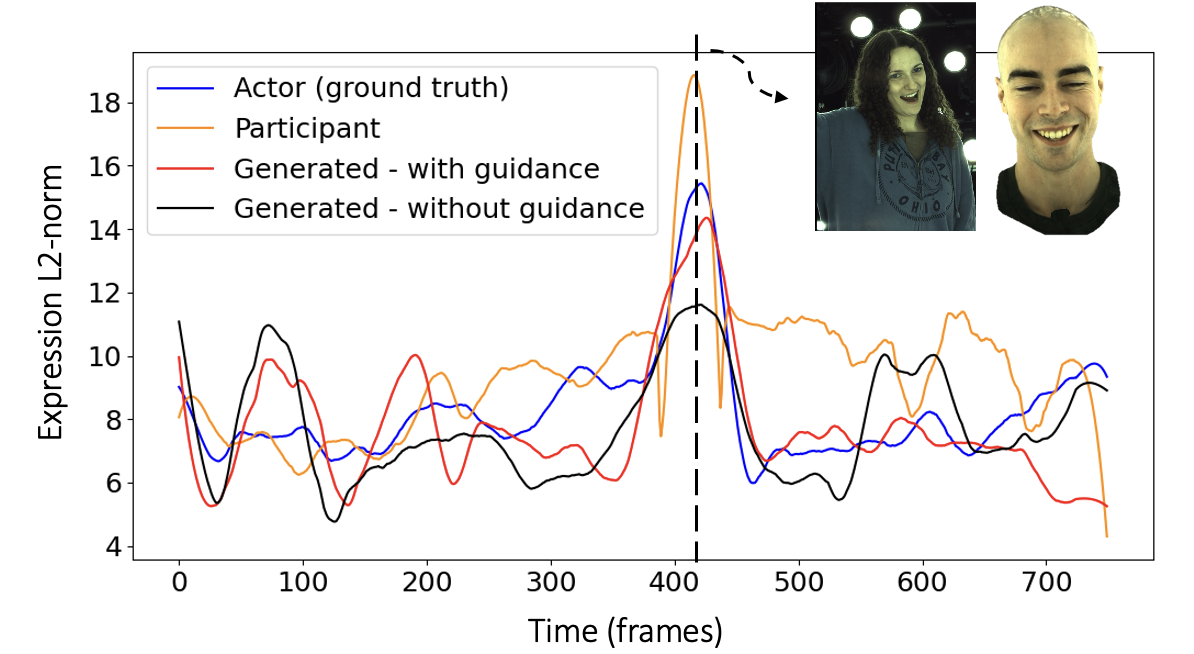}
   \caption{\textbf{Guidance over time}. L2 norm of expression codes for the ground truth actor, the participant, and the generated actor with or without guidance. With guidance, \MethodName produces the appropriate reaction in dyadic interactions.}
   \label{fig:plot_guidance}
   \vspace{-10pt}
\end{figure}

\subsection{Audio-Visual Guidance in Conversations}

As mentioned in \cref{sec:method_dyadic}, we propose to provide audio-visual guidance from a participant in dyadic conversations, extending \MethodName to \textit{conversational} avatars. Very few methods put avatars in this setting, actively talking and listening, although this interaction is common in everyday life. Audio2Photoreal~\cite{ng2024audio2photoreal} produces listening behavior based on audio input, but cannot react to participant's expressions.

\cref{fig:qual_smirk} shows the same actor talking with different participants. The basic \MethodName (without any guidance) is shown in the last column. With our audio-visual guidance (3rd col.), the generated avatar reacts with their gaze and/or smile, according to the participant's expression or voice.  
\cref{fig:plot_guidance} shows the L2 norm of the SMIRK~\cite{SMIRK:CVPR:2024} expression codes over time for the ground truth actor, participant, and generated actor with and without guidance. The graph shows how the actor reacts at the same time or before/after the participant, while they interact. With the proposed guidance, \MethodName produces the appropriate expression, leading to empathetic interactions. We also noticed a $2\%$ decrease in FD$_{e}$ for our test set with this guidance (see also suppl.).

% \textbf{Figures}

% - Figure with comparisons with other methods (show lip-syncing and corresponding phoneme)

% - Figure showing visual guidance from participant (expression and reaction), listening behavior, nodding, smiling/laughing, surprise

% - Figure showing diverse expressions for same phoneme 

% - Figure showing beats and corresponding head motion / expression

% - Figure showing expressions of both in time (showing correlation / reactions) ?

% \textbf{Comparisons with audio-driven methods for talking faces:}

% - VASA-1: our DiT model without flow matching is similar with VASA-1. It is identity-specific and we use our own decoder.

% - Sadtalker ? (audio to expression mapping)

% - DreamTalk: diffusion-based - similar

% - Diffused Heads

% - DiffTalk

% - Evonne's photoreal

% FaceTalk~\cite{aneja2023facetalk}: diffusion with wav2vec

% Should we add methods like FaceFormer, CodeTalker, ScanTalk?

% \textbf{Comparisons with text-driven methods for talking faces:}

% - Faces that Speak~\cite{Jang_2024_CVPR}: GAN-based + lip-sync loss + flow matching in the middle for generating motion features. They generate both speech and video from text. They don't have code ? We can compare with their fusion - simple addition ?

% - We can compare with audio-driven methods (e.g. top 2), by using as input our generated speech (cascaded approach: TTS + audio-driven method)

\section{Conclusion}
\label{sec:conclusion}

In conclusion, we introduce a novel method for joint audio-visual generation of 4D talking avatars, given only text inputs (\eg, obtained by an LLM). Our fully-parallel diffusion-based architecture ensures cross-modal communication, leading to synchronized audio and visual modalities. Trained with flow matching, \MethodName leads to fast inference and natural-looking talking faces. It also enables dyadic conversations, animating an always-on avatar that actively listens and reacts to the audio-visual input of a user. We believe that this work gets one step closer to enabling natural interaction between a human and an AI system. 

\noindent
\textbf{Limitations and Ethical Considerations.}
While our method produces natural-looking avatars, it does not have any understanding of the semantic content of the inputs. E.g., if a user makes a joke without laughing themselves, the model would not know that it was a joke and could not react accordingly. This will be an interesting exploration for future work. In this work, we only use consenting participants. 
Since our method is identity-specific, only these can be rendered. 
This addresses ethical concerns of generating non-consenting subjects.

{
    \small
    \bibliographystyle{ieeenat_fullname}
    \bibliography{main}
}

\clearpage
\setcounter{page}{1}
\maketitlesupplementary

\renewcommand\thesection{\Alph{section}}

\setcounter{section}{0}

\section*{Contents}

\noindent
The supplementary document is organized as follows: 
\begin{enumerate}
    \item[A.] Additional Results
    \item[B.] Implementation Details
    \item[C.] Ethical Considerations
    % \item Limitations
    % \item Ethical Considerations
\end{enumerate}
We strongly encourage the readers to watch our supplementary video.

\section{Additional Results}

\begin{figure}[t]
  \centering
   \includegraphics[width=\linewidth]{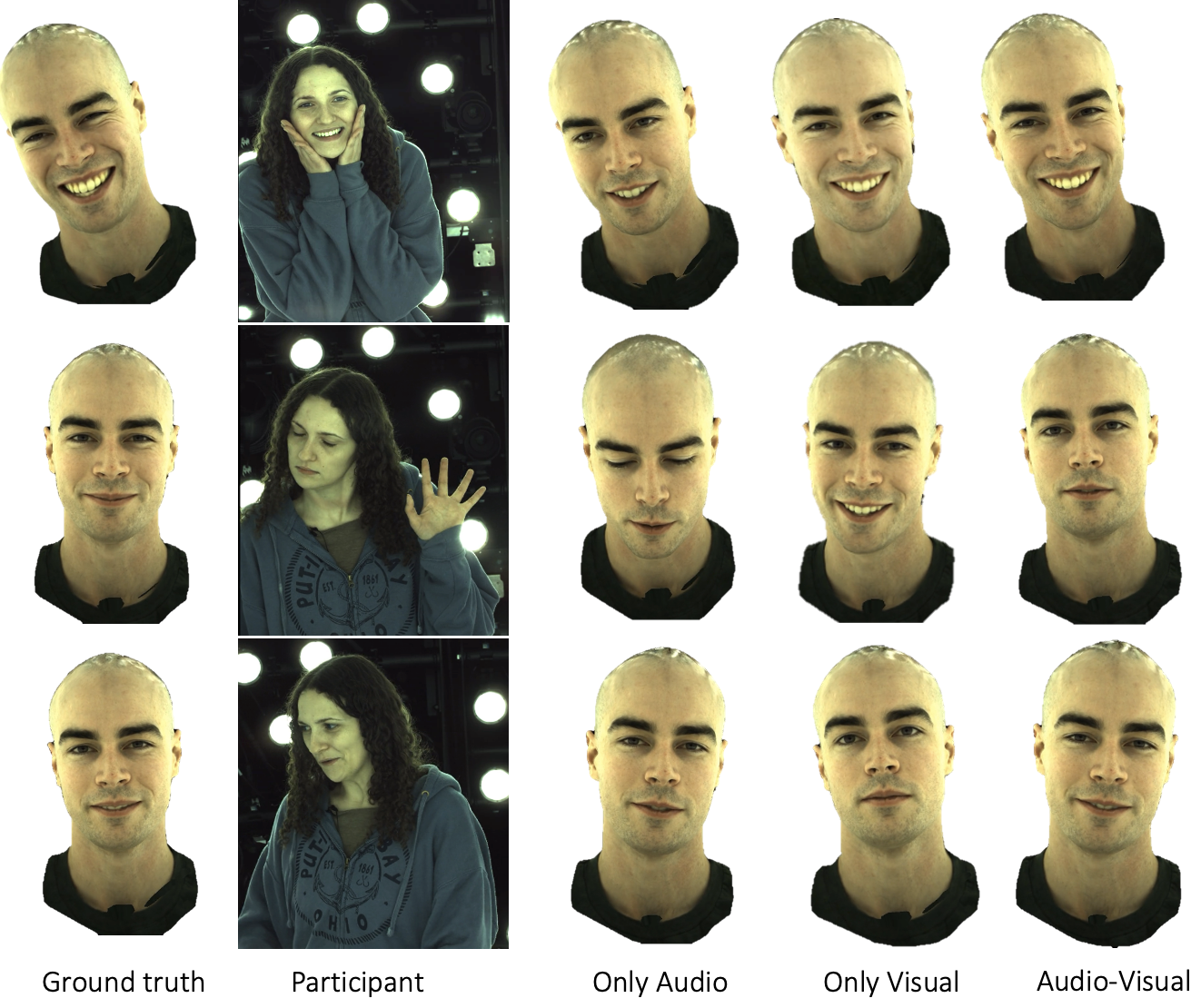}
   \caption{\textbf{Audio-Visual Guidance.} 
   \MethodName with audio-visual guidance produces more realistic expressions while listening. In audio-only guidance, the avatar might lose eye contact or not mirror a smile. In visual-only guidance, it might produce more smiles than needed.}
\label{fig:audioonly_visiononly_audiovisual}
\end{figure}

\noindent
\textbf{Audio-Visual Guidance.}
As mentioned in~\cref{sec:method_dyadic}, we propose to provide audio-visual guidance from a participant in dyadic conversations. We condition the model to visual information, by extracting features $\bm{s}_i$ using SMIRK~\cite{SMIRK:CVPR:2024} from the monocular video of the participant. For the audio, we further extract ASR tokens $\bm{a}_i^{p}$ from the audio channel of the participant. We notice that both modalities are important, in order to produce realistic and meaningful interactions. However, overall for our conversational data, the results feel realistic even if we condition on only one modality. In~\cref{fig:audioonly_visiononly_audiovisual}, we demonstrate some cases where we notice some difference when only one modality is available. In the first row, the avatar better mirrors the expression, with a wider smile, when visual information is present. In the second row, it produces a realistic  but unnecessary smile with only visual guidance, whereas it loses eye contact with only audio guidance. In the third row, the audio seems to play an important role that makes the user to smile. Overall, since we have available both audio and video, we propose to condition \MethodName to both modalities to produce our photo-realistic \emph{always-on} avatar.

% As mentioned in the Sec.~4.3, we noticed a $2\%$ decrease in FD$_{e}$ for our test set of dyadic conversations, by adding the proposed audio-visual guidance.
\cref{tab:quant_with_without_smirk} shows the quantitative results for the basic \MethodName without guidance and with audio-visual guidance. We compute the F1-score for the lip closures, as well as the F1-score for the smiles. The lip closures are detected by measuring the distance of the vertices of the inner upper and lower lips for the 3D mesh per frame. Similarly, smiling is detected when the distance of the left and right corners of the mouth is larger than a threshold. We also compute the Fr\'echet distance between ground truth and generated face expressions (FD$_{e}$) to estimate the distribution distance. With our proposed guidance, we notice an increase in F1-score for the smiles and a decrease in FD$_{e}$ for the dyadic setting, as the avatar produces more realistic reactions and facial expressions while listening to the user. Our basic \MethodName achieves a slightly more accurate lip synchronization.

% \begin{table*}[t]
%   \centering
%   \begin{tabular}{@{}l|ccccccccc@{}}
%   \toprule
%     & \textit{Lip Sync} & \multicolumn{1}{c}{\textit{Realism}} & \multicolumn{2}{c}{\textit{Diversity}} & \multicolumn{2}{c}{\textit{AV-Alignment}} & \multicolumn{2}{c}{\textit{Audio Quality}} \\
%     \midrule
%     Method & F1$_{lips}$$\uparrow$ & FD$_{e}$$\downarrow$ & Div$_{h}$$\uparrow$ & Div$_{e}$$\uparrow$ & BC$_{h}$$\uparrow$ & BC$_{e}$$\uparrow$ & MCD$\downarrow$ & WER$\downarrow$ \\
%     \midrule
%     AV-Flow w/ Fusion (b) & 0.865 & \textbf{0.828} &  0.022 & \textbf{0.695} & \textbf{0.271} & 0.186 & 0.880 & \\
%     AV-Flow (Ours) & \textbf{0.964} & \textbf{0.861} & \textbf{0.029} & \textbf{0.680} & \textbf{0.258} & \textbf{0.229} & \textbf{0.900} & \textbf{0.157}\\
%     \bottomrule
%   \end{tabular}
%   \caption{\textbf{Ablation Study.} }
%   \label{tab:quant_ablation_2}
%   \vspace{-10pt}
% \end{table*}

\begin{table}[t]
  \centering
  \begin{tabular}{@{}l|ccc@{}}
  \toprule
    % & \textit{Lip Sync} & \multicolumn{1}{c}{\textit{Realism}} & \multicolumn{2}{c}{\textit{Diversity}} & \multicolumn{2}{c}{\textit{AV-Alignment}} & \multicolumn{2}{c}{\textit{Audio Quality}} \\
    % \midrule
    Method & F1$_{lips}$$\uparrow$ & F1$_{smiles}$$\uparrow$ & FD$_{e}$$\downarrow$ \\
    \midrule
    AV-Flow w/o guidance & \textbf{0.964} & 0.611 &  0.861 \\
    AV-Flow w/ guidance & 0.933 & \textbf{0.685} & \textbf{0.845} \\
    \bottomrule
  \end{tabular}
  \caption{\textbf{AV-Flow with or without Guidance.} In dyadic conversations, the proposed audio-visual guidance leads to more realistic reactions and facial expressions, while the avatar is listening to the user. Without guidance, our basic \MethodName achieves slightly more accurate lip synchronization.}
  \label{tab:quant_with_without_smirk}
  % \vspace{-10pt}
\end{table}

\begin{figure}[t]
  \centering
   \includegraphics[width=\linewidth]{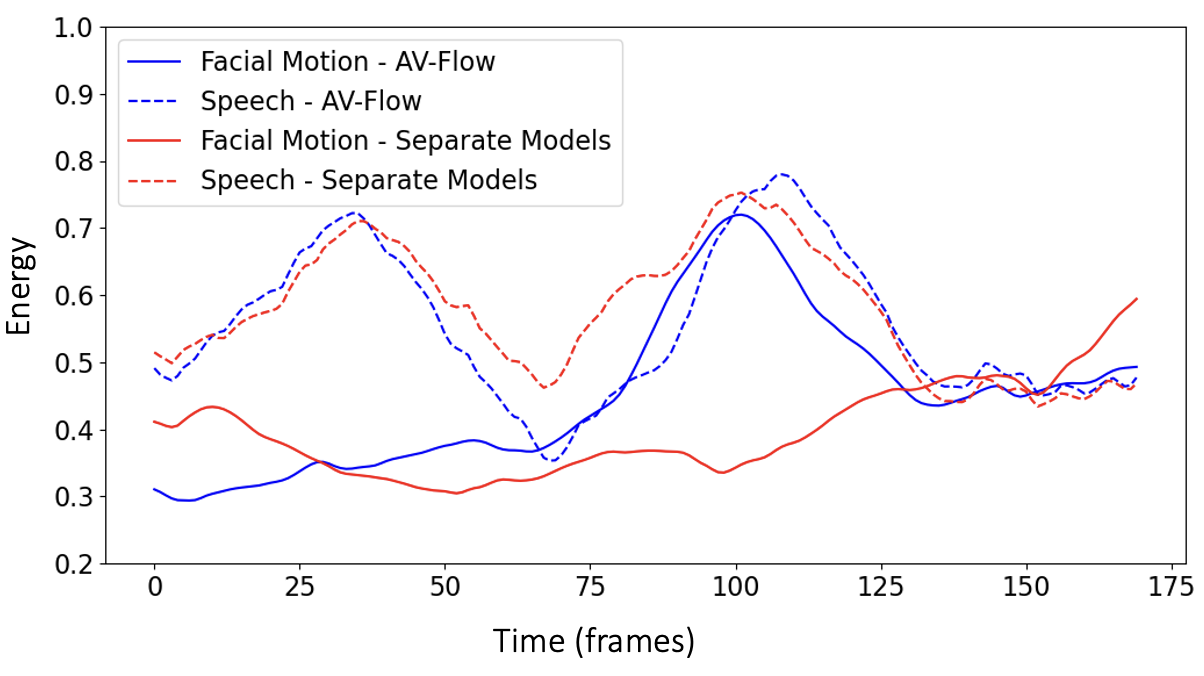}
   \caption{\textbf{Audio-Visual Alignment.} Correlation between synthesized speech and facial motion by AV-Flow, compared to the variant of separate models without any connections. Energy is estimated as the normalized squared L2-norm of the generated facial dynamics and mel-spectrogram over time.}
   \label{fig:audio_exp_correlation}
\end{figure}

\noindent
\textbf{Audio-Visual Alignment.}
We design intermediate highway connections that enable communication between the audio and visual diffusion transformers.
In our ablation study in~\cref{sec:exp_ablation}, we notice that our proposed audio-visual fusion leads to the best correlation between audio and motion, as measured by the BC$_{h}$ and BC$_{e}$ metrics.
This audio-visual correlation is also shown in \cref{fig:audio_exp_correlation}. We observe patterns where the energy of the facial motion matches the energy of the corresponding synthesized audio (plotted as the normalized squared L2-norm of the generated facial dynamics and mel-spectrogram over time). We compare with the variant of the separate models, without any cross-modal connections, where the correlation is lower.

\begin{figure*}[t]
  \centering
   \includegraphics[width=0.9\linewidth]{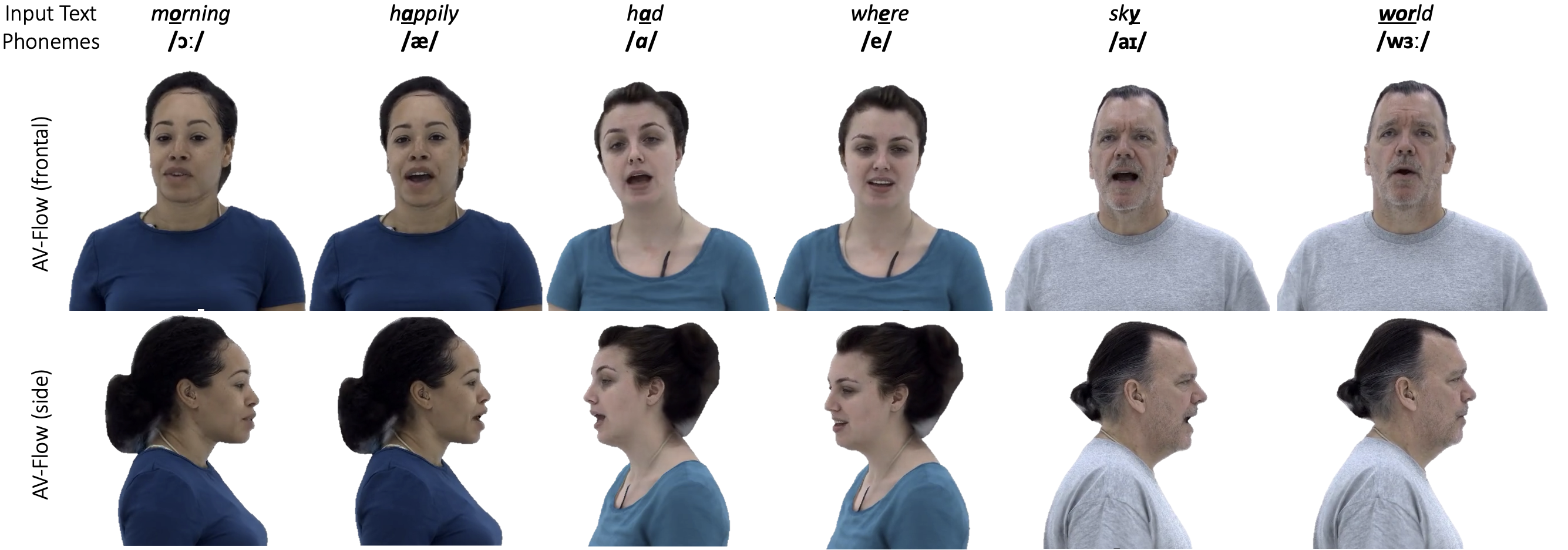}
   \caption{\textbf{Qualitative Results of \MethodName.} We show frontal and side views for corresponding phonemes. We use pre-trained personalized renderers~\cite{timur2021driving_renderer} that synthesize photo-realistic 3D avatars.}
   \label{fig:frontal_side_views}
\end{figure*}

\noindent
\textbf{Additional Qualitative Results.}
\cref{fig:frontal_side_views} shows additional qualitative results of our method, rendered in frontal and side views. We use pre-trained personalized renderers~\cite{timur2021driving_renderer,ng2024audio2photoreal} that produce photo-realistic 3D avatars. Therefore, we can render the generated avatars from any viewpoint. In this work, we are mostly interested in the facial expression, lip synchronization and realistic head motion over time. We only provide the side views for completeness. We refer the interested reader to~\cite{timur2021driving_renderer,deep_appearance} for more details in Codec Avatars.

\noindent
\textbf{Video Results.}
We encourage the readers to watch our supplementary video.

\section{Implementation Details}

\noindent
\textbf{Text-to-Tokens.} 
Since our training dataset does not include any text annotations, we extract tokens (logits) from the raw audio using an ASR model (see~\cref{sec:method_representations}). At inference time, in order to be able to synthesize audio-visual content directly from text characters, we learn a text-to-tokens model. Inspired by the architecture proposed by Matcha-TTS~\cite{mehta2024matcha}, we first map the input text to phonemes. We learn phoneme embeddings (192-dimensional) that are passed through a text encoder of 3 1D convolutional layers. A duration predictor gives their duration. A diffusion transformer of 3 layers maps the features to logits, which can be used as input tokens to our model. We follow the rest hyperparameters, architecture and training with flow matching of Matcha-TTS~\cite{mehta2024matcha}. Our main difference is that we predict character-level logits, not mel-spectrograms. We use LJSpeech~\cite{ljspeech17} to train our text-to-tokens model.

\noindent
\textbf{Architecture.}
We use $N=8$ blocks for our audio and visual DiTs. We first project the inputs to 512-dimensional through a linear layer. Each transformer block has input and output dimensions of 512, hidden size of 1024, and 4 heads for the multi-head self-attention.
We use windows of 10 frames, looking only 2 frames in the future. With this windowing, we achieve only 120ms latency, as mentioned in~\cref{sec:experiments}. We have also tried windows of 20 frames, getting similar results, but a bit higher latency.
As described in~\cref{sec:method_architecture}, we upsample the data at 86fps to achieve exact correspondence between audio and video. We extract mel-spectrograms following the same extraction as BigVGAN~\cite{lee2022bigvgan}. In this way, we directly use the pre-trained BigVGAN as our vocoder to get the output speech signal. Our input tokens are extracted from a pre-trained Wav2Vec2 model with the base architecture, that is trained for ASR using 960 hours of unlabeled audio from the LibriSpeech dataset~\cite{paszke2019pytorch,panayotov2015librispeech,yang2021torchaudio}.

\noindent
\textbf{Training.}
During training, we use a batch size of 16 segments. Each segment corresponds to a duration of 20 seconds.
We set $\sigma_\text{min} = 10^{-6}$.
Our implementation is based on PyTorch~\cite{paszke2019pytorch}. We use AdamW optimizer~\cite{loshchilov2017decoupled} with a learning rate of $10^{-4}$, and hyperparameters $\beta_1 = 0.9$, $\beta_2 = 0.98$, $\epsilon = 10^{-9}$. We train \MethodName for about 36 hours (1 million iterations) on a single A100 GPU.

\section{Ethical Considerations}

We use the publicly available dataset proposed by Audio2Photoreal~\cite{ng2024audio2photoreal}. We also collected an additional dataset of 50 hours in a similar setting. Both datasets include dyadic conversations between individuals. The data include raw audio and video, as well as face expression codes and head poses for the actors, paired with pre-trained personalized renderers~\cite{timur2021driving_renderer,deep_appearance}. During collection of the data, we have followed appropriate procedures and all individuals have provided their full consent for our research work. Our model is identity-specific and thus, only those individuals can be rendered and no one else. This addresses ethical concerns of generating subjects without their consent, or generating misleading content. We have also used audio from the EARS dataset~\cite{richter2024ears} to test lip synchronization to custom input speech and the widely used LJSpeech~\cite{ljspeech17} to train our text-to-tokens module.

Although we have strictly followed all these procedures in collecting and using our data, we would like to note the potential misuse of similar technologies in generating photo-realistic human avatars. Apart from the benefits for education, virtual communication, healthcare, etc, there is still the possibility of generating misleading content. Research on fake content detection and forensics is crucial.
We intend to release our source code to help improving such research.

% \section{Rationale}
% \label{sec:rationale}
% % 
% Having the supplementary compiled together with the main paper means that:
% % 
% \begin{itemize}
% \item The supplementary can back-reference sections of the main paper, for example, we can refer to \cref{sec:intro};
% \item The main paper can forward reference sub-sections within the supplementary explicitly (e.g. referring to a particular experiment); 
% \item When submitted to arXiv, the supplementary will already included at the end of the paper.
% \end{itemize}
% % 
% To split the supplementary pages from the main paper, you can use \href{https://support.apple.com/en-ca/guide/preview/prvw11793/mac#:~:text=Delete%20a%20page%20from%20a,or%20choose%20Edit%20%3E%20Delete).}{Preview (on macOS)}, \href{https://www.adobe.com/acrobat/how-to/delete-pages-from-pdf.html#:~:text=Choose%20%E2%80%9CTools%E2%80%9D%20%3E%20%E2%80%9COrganize,or%20pages%20from%20the%20file.}{Adobe Acrobat} (on all OSs), as well as \href{https://superuser.com/questions/517986/is-it-possible-to-delete-some-pages-of-a-pdf-document}{command line tools}.

\end{document}